\documentclass[10pt,twocolumn,letterpaper]{article}

\usepackage{iccv}              

\definecolor{iccvblue}{rgb}{0.21,0.49,0.74}
\usepackage[pagebackref,breaklinks,colorlinks,allcolors=iccvblue]{hyperref}

\usepackage[accsupp]{axessibility}
\usepackage{multirow}
\usepackage{hyperref}

\usepackage{balance}
\usepackage{lastpage}

\usepackage{algorithm}
\usepackage{algpseudocode}

\hypersetup{
    colorlinks=true,
    citecolor=green,
    linkcolor=red,
    filecolor=brown,      
    urlcolor=blue,
    pdfpagemode=FullScreen,
    }
    
\def\paperID{15989} 
\def\confName{ICCV}
\def\confYear{2025}

\title{PROL : Rehearsal Free Continual Learning in Streaming Data via Prompt Online Learning}
\author{M. Anwar Ma'sum$^{1,*}$, Mahardhika Pratama$^{1}$, Savitha Ramasamy$^{2}$, Lin Liu$^{1}$, \\ Habibullah Habibullah$^{1}$, and Ryszard Kowalczyk$^{1,3}$\\
\small $^1$STEM University of South Australia, $^2$Institute for Infocomm Research, A*STAR \& IPAL, CNRS@CREATE \\
\small $^3$Systems Research Institute, Polish Academy of Sciences, Warsaw, Poland 
}

\begin{document}

\newcommand\blfootnote[1]{%
  \begingroup
  \renewcommand\thefootnote{}\footnote{#1}%
  \addtocounter{footnote}{-1}%
  \endgroup
}

\maketitle
\begin{abstract}
The data privacy constraint in online continual learning (OCL), where the data can be seen only once, complicates the catastrophic forgetting problem in streaming data. A common approach applied by the current SOTAs in OCL is with the use of memory saving exemplars or features from previous classes to be replayed in the current task. On the other hand, the prompt-based approach performs excellently in continual learning but with the cost of a growing number of trainable parameters. The first approach may not be applicable in practice due to data openness policy, while the second approach has the issue of throughput associated with the streaming data. In this study, we propose a novel prompt-based method for online continual learning that includes 4 main components: (1) single light-weight prompt generator as a general knowledge, (2) trainable scaler-and-shifter as specific knowledge, (3) pre-trained model (PTM) generalization preserving, and (4) hard-soft updates mechanism. Our proposed method achieves significantly higher performance than the current SOTAs in CIFAR100, ImageNet-R, ImageNet-A, and CUB dataset. Our complexity analysis shows that our method requires a relatively smaller number of parameters and achieves moderate training time, inference time, and throughput. For further study, the source code of our method is available at \url{https://github.com/anwarmaxsum/PROL}.
\end{abstract}
\blfootnote{*Corresponding: masmy039@mymail.unisa.edu.au}
\vspace{-9pt}
\section{Introduction}
\label{sec:intro}
Continual learning (CL) has becomes a spotlight in the world of artificial intelligence (AI) involving dynamic environments, as it tackles catastrophic forgetting for a continually evolving sequence of tasks \citep{CLSURVEY1_de2021continual,CLSURVEY2_wang2024comprehensive}. CL methods are proven to be promising in various domains, e.g., computer vision, automation, NLP and graph analysis \citep{CLCV_liu2023incremental,CLNLP_biesialska2020continual,CLGRAPH_tian2024continual,CLAUTO_shaheen2022continual}. Online continual learning (OCL) is an advanced sub-problem of CL where CL is conducted in streaming data. The limitation of seeing data only once in OCL leads to insufficient learning of a deep learning model that results in difficulty in gaining knowledge of a new task while maintaining knowledge of previously learned tasks. In other words, it complicates how to resolve the catastrophic forgetting problem. 
To handle such a limitation, most of the state-of-the-art OCL methods (SOTAs) save previous task exemplars for a rehearsal process to maintain the knowledge from previously learned tasks \citep{ALF_seo2025budgeted,ESRM,LPR_ICML2024}. Practically, this approach is not reliable since, due to privacy constraints or data openness policy, previously learned data can be unavailable anymore. In addition, the rehearsal process increases memory consumption and computational costs.

Leveraging the recent advancement of the foundation model, recent SOTAs utilized a pre-trained model (PTM) that carries a generalization attribute as initial weight \cite{NSCE_xinrui2025forgetting,RANDUMB_prabhu2025randumb,RANPAC_mcdonnell2023ranpac}. However, the SOTAs still store and replay previously learned exemplars or features in their memory. The insight tells us that rehearsal-free online continual learning remains an open problem. On the other hand, the prompt-based approach has been proven to be effective in rehearsal-free continual learning. Specifically, the growing prompt components methods \citep{CONVP_roy2024convolutional,EVO_kurniawan2024evolving,CODA_smith2023coda} outperform the prompt-pool method \citep{L2P_wang2022learning} and task-wise prompt methods \citep{DualP_wang2022dualprompt, HIDE_wang2024hierarchical, CPROMPT_gao2024consistent, SPROMPT_wang2022s} both in terms of accuracy and forgetting measure. However, from the perspective of OCL, the growing components prompt methods violate the throughput aspects, as the model parameters grow with the increasing number of classes. This growth incurs more operations and longer training time, leading to more skipped (unprocessed) data streams. Thus, the approach is not favorable for dealing with catastrophic forgetting in streaming data.

These intriguing insights motivate us to develop a novel rehearsal-free method that can effectively address the OCL problem. We develop our proposed method based on two fundamental goals in OCL, i.e., (1) achieving stability-plasticity and (2) being efficient both in the training and inference phases. To achieve these goals, we design our method with the following components: (1) a single lightweight prompt generator for all tasks, (2) a set of scalers and shifters associated with the class-wise learnable keys, (3) preservation of PTM generalization via cross-correlation matrices, and (4) hard-soft learning updates for a better parameter tuning. The generator is trained only on the first task to avoid forgetting and achieve stability, while scalers and shifters are trained on every task to achieve plasticity. To achieve high throughput, our lightweight generator has fewer parameters than a single generator in \citep{CONVP_roy2024convolutional} and fewer than a single task-specific prompt in \citep{DualP_wang2022dualprompt}. The third component is adapted from SAFE\cite{SAFE_zhao2025safe} to the OCL problem.

In this study, our contributions are: (1) We highlight rehearsal-free constraint in online continual learning and propose a novel prompt online learning (PROL) that addresses the two essential goals of OCL. (2). We design a new mechanism to generate prompts by a single generator with a few smaller-shifter parameters. We also propose a joint loss function for our method. (3). Our rigorous experiment and analysis prove the superiority of our method compared to the existing SOTAs in general and historical performance. Our analysis shows that our method has relatively low parameters, moderate training, inference time, and throughput compared to the SOTAs. (4). We provide a complexity analysis of our method and complete numerical results for further study; please see the supplementary.


\section{Literature Review}
\label{sec:literature}
\noindent \textbf{Online Continual Learning (OCL):} Recent OCL studies shed light on various approaches to handle catastrophic forgetting in streaming data. The backbone tuning approach tunes the whole model with unique mechanisms, e.g.  cross-task class discrimination as in GSA \citep{GSA_guo2023dealing}, dual-view as in DVC \citep{DVC_gu2022not}, proximal point method as in LPR \citep{LPR_ICML2024}, real-synthetic similarity maximization as in ESRM \citep{ESRM}, and adaptive layer freezing as in ALF \citep{ALF_seo2025budgeted}. The other approaches tune the class prototype along with the backbone as in ONPRO \citep{ONPRO_wei2023online}, augmenting and refining the prototype as in DSR \citep{DSR_huo2024non}, or generating multiple features in angular space as in EARL \citep{EARL_seo2024learning}. Another unique approach uses a student-teacher model and knowledge distillation between them, as in MKD \citep{MKD_ICML2024}, CCLDC \citep{CCLDC_wang2024improving}, and MOSE \citep{MOSE_wang2024improving}. The recent development of the foundation model gives rise to the PTM-based OCL, e.g., NSCE \citep{NSCE_xinrui2025forgetting} that transfers PTM weight and fine-tunes it, RandDumb \citep{RANDUMB_prabhu2025randumb} that trains RBF kernel from PTM generated features, and RanPAC \cite{RANPAC_mcdonnell2023ranpac} that projects the feature with a randomized matrix. Tackling OCL can be achieved by a customized optimization technique, e.g., POCL \citep{POCL_wu2024mitigating} that utilizes Pareto optimization, and CAMA \citep{CAMA_kim2024online} that performs confidence-aware moving average. However, these aforementioned methods still save exemplar or features for rehearsal. It is not always applicable in real-world applications since the old data may not be available or accessible. Note that rehearsal-based CL methods \cite{ICARL_rebuffi2017icarl,BIC_wu2019large,GDUMB_prabhu2020gdumb,DER_buzzega2020dark,XDER_boschini2022class} work poorly in OCL and even worse than the aforementioned methods.

\noindent \textbf{PTM-Based Continual Learning:}  The prompt-based approach is one of the most popular approaches in CL due to its simplicity and scalability. The prompt can be generated by learnable latent parameters or a vector generated by trainable generator networks and then prepended into visual transformer (ViT) multi head self attention  (MSA) \citep{PROMPTUN_lester2021power,PREFIXTUN_li2021prefix}. The prompt structure can be organized by a pooling approach as in L2P \citep{L2P_wang2022learning}, task-specific and task-shared structure as in DualPrompt, S-Prompt, HiDePrompt and CPrompt \citep{DualP_wang2022dualprompt,SPROMPT_wang2022s,HIDE_wang2024hierarchical,CPROMPT_gao2024consistent}, evolving prompt component as in CODA-P\cite{CODA_smith2023coda}, evolving prompt generator as in ConvPrompt and EvoPrompt \cite{CONVP_roy2024convolutional,EVO_kurniawan2024evolving}. The other approach utilizes Low-Rank Adaptation (LoRA) where it extend the frozen ViT weight $W$ with learnable matrices $A$ and $B$ \cite{LAE_gao2023unified, CLORA_smith2023continual,InfLORA_liang2024inflora}. The adapter-based approach emerges as a new PTM approach for CL by extending the ViT with a sequence of (growing) adapters \citep{EASE_zhou2024expandable,MOS_sun2024mos}. Rather than extending the inner level, such as ViT weight, adapters work on the feature level by projecting the ViT feature into a more discriminative form.  Last but not least, recent works focused on learning mechanisms, i.e., slow-learner rather than structures to elevate PTM model discrimination \citep{SLCA_zhang2023slca, SAFE_zhao2025safe}. Except for the adapters approach, all the mentioned methods were developed for rehearsal-free CL. Even thought they have been proven effective in rehearsal-free CL, the methods are not yet proven in rehearsal free OCL where the data is streaming and can be seen only once.   

\begin{figure*}[h!]
\setlength{\abovecaptionskip}{-7pt plus 0pt minus 0pt}
\centering
\begin{center}
\includegraphics[width=1.0\textwidth]{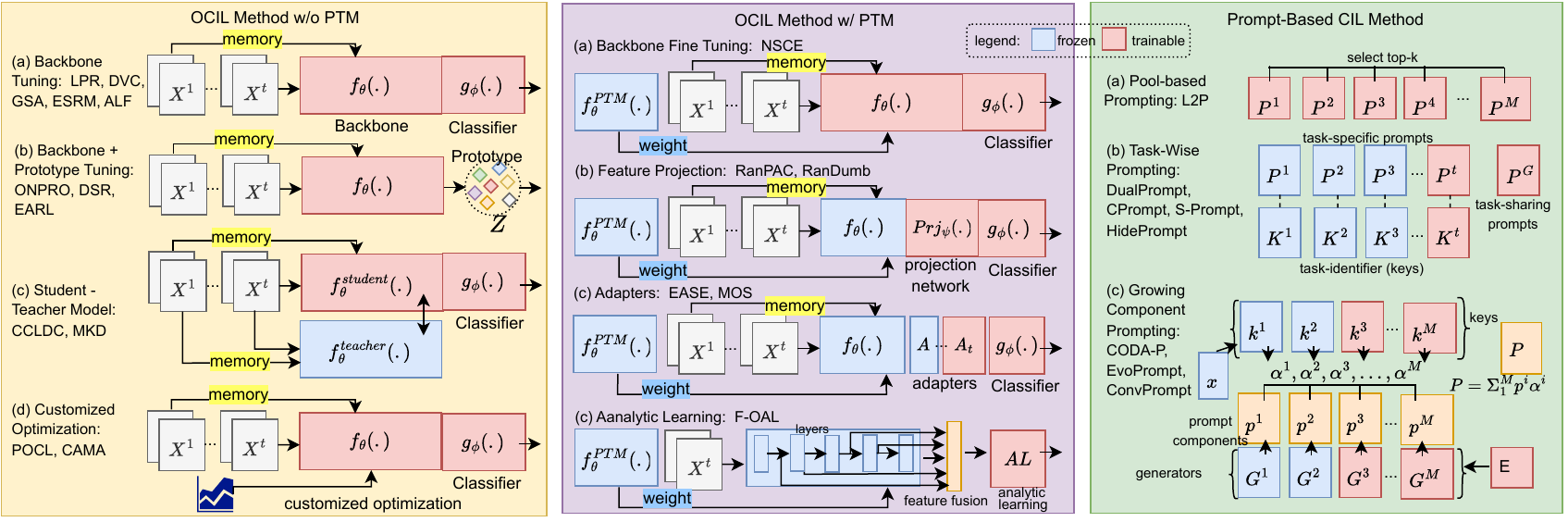}
\end{center}
\caption{Topology of OCL and prompt-based CL methods (a) OCL Methods without pre-trained model (PTM), (b) OCL methods with PTM as weight initiation, (c) Prompt-based CL method. Majority of the OCL method requires memory from previously learned classes to be replayed along with the currently learned classes, while the prompt-based method is rehearsal-free.} 
\label{fig:ocl_topology}
\end{figure*}

\section{Preliminary}
\label{sec:preliminary}
\textbf{(a) Problem Formulation :} Class incremental learning (CIL) problem is defined as the problem of learning a sequence of fully supervised tasks $\{\mathcal{T}^t\}_{t=1}^T$, where after finishing learning a task $\mathcal{T}^t$, a model must recognize all learned tasks i.e. $\mathcal{T}^1$,$\mathcal{T}^2$,...$\mathcal{T}^{t-1}$,$\mathcal{T}^t$. Symbol $T$ represents the number of consecutive tasks. Each task carries pairs of training samples, i.e., $\mathcal{T}^{t}=\{(x_i^{t},y_i^{t})\}_{i=1}^{|\mathcal{T}^t|}$ where $x_i\in\mathcal{X}^t$ and $y_i\in\mathcal{Y}^t$ denotes input image and its label, while $|.|$ denotes cardinality. $\mathcal{C}^{t}$ denotes the unique class labels in $\mathcal{Y}^t$, i.e., $\mathcal{C}^{t}=unique(\mathcal{Y}^{t})$, and $|\mathcal{C}^{t}|$ denotes the number of classes in $\mathcal{T}^t$. The task ID is visible only during the training phase of CIL, unlike in task incremental learning (TIL), which is visible in both the training and inference phases.

Online Class incremental learning (OCIL) is similar to CIL but the data comes in streams $s_1, s_2, s_3, ...$ where $s_j=\{(x_i^{t},y_i^{t})\}_{i=1}^{|s_j|}$ rather than in a storable $\mathcal{T}^{t}=\{(x_i^{t},y_i^{t})\}_{i=1}^{|\mathcal{T}^t|}$ manner.  $|s_j|$ is relatively small, e.g., 10. The data can be seen only once, as the data will not be available (stored) after streaming. Following \citep{iBLURRY_koh2021online}, there are three main settings in OCL, i.e., disjoint task where each task $t$ is disjoint from another task $t'$, i.e., $\forall t, t'\neq t, (\mathcal{T}^t\cap \mathcal{T}^{t'}=\emptyset$), blurry task setting where there is no explicit boundary between tasks, and iBlurry setting where consist of $N\%$ of the classes are put into disjoint tasks and the other $(100-N)\%$ classes appear in every task. In this study, we focus on the disjoint task setting that is the most common in OCL and CL.

\noindent \textbf{(b) The OCL Method and Its Gap:} Figure \ref{fig:ocl_topology} shows a topology that the OCL methods can be divided into 2 main categories, i.e., OCL method without PTM (Non-PTM) and with PTM. In the perspective of available foundation model, the main gap of this category is that it tunes the whole backbone $f_{\theta}(.)$ and classifier $g_{\phi}(.)$ from scratch in the case of streaming data where the data can be seen only once. In addition, the backbone training in every task increases the chance of the model forgetting.  Even with the presence of sample memory to avoid forgetting, the methods can not perform well since its weight is not yet convergent. A customized backbone tuning technique such as the proximal point method,  real-synthetic similarity maximization, dual view, or adaptive layer freezing \citep{LPR_ICML2024,DVC_gu2022not,GSA_guo2023dealing,ESRM,ALF_seo2025budgeted}, or customized optimization method as in \citep{POCL_wu2024mitigating, CAMA_kim2024online} can not compensate the insufficient training and avoid model fofgetting. The other approach trains dual models, i.e., student ($f_{\theta}^{student}$) and teacher ($f_{\theta}^{teacher}$) backbones. Even though one of the model is frozen, this approach requires higher training complexity (longer training time) and reduces model throughput.  The prototype approach replaces the MLP classifier ($g_{\theta}$) with a prototype-based classifier ($g_{Z}$). Despite reducing model forgetting, the prototypes do not optimally discriminate the learned classes since the prototypes are calibrated with once-seen data.

On the other side, the PTM-based method leverages PTM generalized weight as its initial values. NSCE\citep{NSCE_xinrui2025forgetting} transfers the PTM weight and then fine-tunes its weight along with the classifier to adapt to the streaming data. Despite achieving better performance than non-PTM methods, it still requires memory samples that may not be applicable in real-world applications due to data privacy or openness policy.  Feature projection network, e.g., RanPAC\citep{RANPAC_mcdonnell2023ranpac} and RanDumb\citep{RANDUMB_prabhu2025randumb} transform PTM features into a more discriminative form by a projection network $Prj_{\psi}$. Similarly, this approach requires memory (in feature space) from a previous task to be replayed in the currently learned class. The adapter methods \citep{EASE_zhou2024expandable,MOS_sun2024mos} that were originally for offline CL are considered as baselines in OCL. Aside from utilizing sample memory, this method has a growing number of adapters ($A_1, A_2,...,A_t$) along with the increasing of learned classes. This mechanism is bounded to increase model complexity and reduce model throughput.  F-OAL \citep{FOAL_zhuang2025f} forms a feature fusion from its PTM layers then classify it with analytical learning (AL). Despite working in rehearsal-free setting, the method requires more operation for AL since the least square solver requires higher computation than standard MLP or prototype-based classifier.

\noindent \textbf{(c). Prompt-based CL Methods and its Drawback:}  The pool-based prompting, e.g., L2P\citep{L2P_wang2022learning} selects top-k prompts from the pool $[P^1, P^2,...P^M]$ where $M$ is the pool size. Any prompt $P^m$ in the pool can be trained for any task $t\in[1..T]$. The task-wise prompting, e.g., DualPrompt\citep{DualP_wang2022dualprompt} trains only the corresponding prompt $P^t$ that associated to task-wise key $K^t$ during the training process on task $t$, while the prompts and keys associated with the previous tasks i.e $[P^1, P^2,...P^{(t-1)}]$ and $[K^1, K^2,...K^{(t-1)}]$ are frozen. The growing component approach dynamically increases the number of prompt components, e.g., $p^i, p^{i+1},..p^M$ to learn new classes while maintaining knowledge on previously learned classes. The final prompt is generated by weighted summing the components, i.e., $P=\Sigma_{i=1}^M p^i\alpha^i$ where $\alpha^i$ denotes similarity value. The prompt components $p^t$ can be a learnable latent parameter as in \citep{CODA_smith2023coda} or generated by prompt generator network, i.e., $p^i = G^i(x)$ as in \citep{CONVP_roy2024convolutional}. 

The pool-based approach has a high risk of forgetting since all the prompts $[P^1, P^2,...P^M]$ are possible to be trained in all tasks $t \in [1..T]$. A prompt $P^i$ is optimal for the $t^{th}$ task but is tuned again in the $t+1^{th}$ task. Therefore, it is no longer optimal for $t^{th}$ task and leads to forgetting. Task-specific prompting has the risk of high similarity keys between two different tasks, e.g., $K^t \approx K^{t'}$ for $t \neq t'$, which leads to inaccurate task prediction during inference. 
The growing component prompting adds new components, i.e., $p^i, p^{i+1},..p^M$ that could disrupt the previous components, i.e., $p^1, p^2,..p^{i+1}$ that are already optimal for previous tasks. Similar to the adapters method, this approach could lead to higher model complexity and reduce its throughput due to the growing prompt components or prompt generators.
\begin{figure*}[h!]
\setlength{\abovecaptionskip}{-7pt plus 0pt minus 0pt}
\centering
\begin{center}
\includegraphics[width=1.0\textwidth]{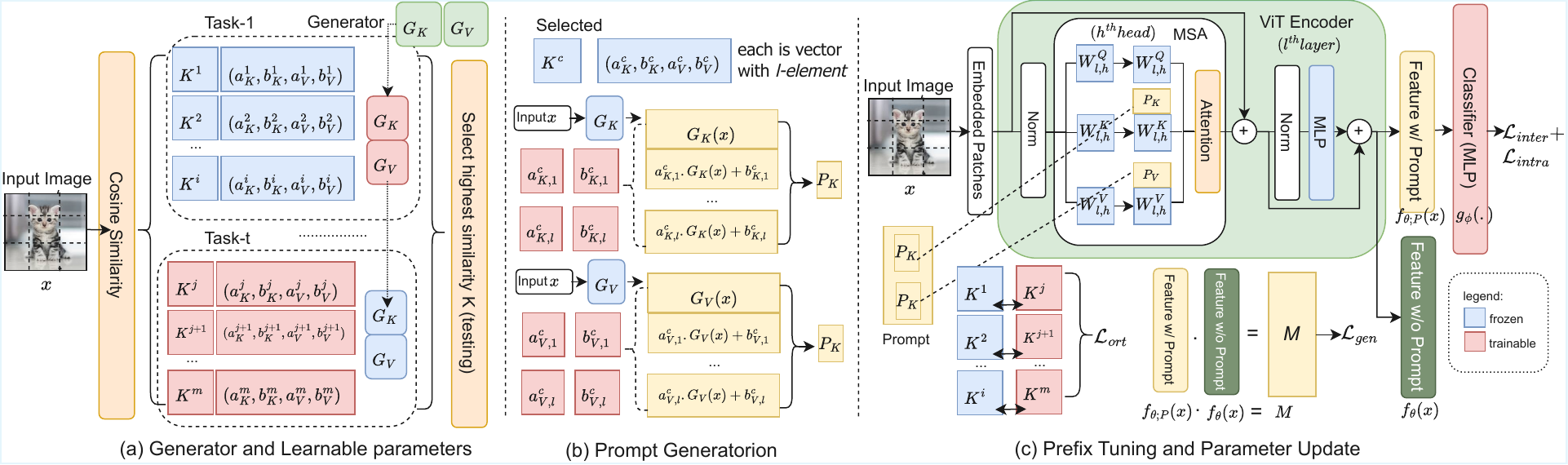}
\end{center}
\caption{\small \textcolor{black}{The proposed PROL diagram. PROL utilizes a single generator $(G_K, G_V)$ for all classes in all tasks, learnable scalers $(a^i_K, a^i_V)$, and shifters $(b^i_K, b^i_V)$ parameters asocciated to the class-wise key $K^i$. The generator is trained only in the first task and frozen afterward, while the scalers, shifters, and class-wise keys are trained in every task}. The prompt is generated by scaling and shifting $G(x)$ by the selected scalers and shifters associated with the highest similarity key. The model is updated by joining the loss function consisting of intra-task loss $\mathcal{L}_{intra}$, inter-task loss $\mathcal{L}_{intra}$, orthogonality loss  $\mathcal{L}_{ort}$, input-to-key similarity loss  $\mathcal{L}_{sim}$ and  generalization loss  $\mathcal{L}_{gen}$.}
\label{fig:prol_fiagram}
\end{figure*}

\section{Proposed Method}\label{sec:method}
In this study, we propose prompt online learning (PROL) to handle catastrophic forgetting in streaming data. The goal of the method is to achieve stability-and-plasticity and high throughput. Thus, we utilize only a single lightweight generator for all classes in all tasks. The generator is trained only in the base task, then frozen afterwards to maintain stability, and assisted by learnable scalers and shifters to achieve plasticity. Figure \ref{fig:prol_fiagram} shows the architecture and flow of PROL, i.e., part \ref{fig:prol_fiagram}(a) shows the generator and learnable parameters, part \ref{fig:prol_fiagram}(b)  shows the prompt generation mechanism, and part \ref{fig:prol_fiagram}(c) shows the model forward, loss, and parameter updates. The following subsections explain the details of the components and the learning mechanism. 

\subsection{Learnable Parameters:}\label{sec:params}
\noindent \textbf{(a). Single Lightweight Generator for All:}  We design a lightweight generator $(G_K, G_V)$ to achieve the highest possible throughput in online continual learning that is trained only in the first task yet can be utilized in all tasks. We design it in a pair following the prompt structure in prefix tuning \citep{PREFIXTUN_li2021prefix} that divides prompt $P$ into two parts, i.e., $P_K$ and $P_V$ that prepend into $K$ and $V$ of the ViT layer's MSA, respectively. Following ConvPrompt \citep{CONVP_roy2024convolutional}, the generator is distributed into all MSA heads of the ViT layers. Thus, the generator can be noted as $(G_K, G_V) = \{(G_{K,l,h}, G_{V,l,h})\}, l \in [1..L] h \in [1..H]$, where $L$ and $H$ are the number of layers and MSA heads per layer respectively. Each $G_{K,l,h}$ or $G_{V,l,h}$ is a 1DCNN kernel with the size of 3, thus only saving 3 parameters. Therefore, the total parameters of our generator are only $2 \times H \times L \times 3$, which is generally less than 1K.

\noindent \textbf{(b). Learnable Scalers-and-Shifters and Class-wise Key:} To assist the generator in achieving plasticity, we train a learnable scalers-and-shifters $(a^c_K, b^c_K, a^c_V, b^c_V)$ associated with a learnable class-wise key $K^c$ (figure \ref{fig:prol_fiagram}(a)), where $c$ is the class label that is represented by $K^c$. Following the generator pair, we can specify that $(a^c_K, b^c_K)$ and $(a^c_V, b^c_V)$ are the scalers-and-shifters for $G_K$ and $G_V$ respectively. However, the scaller and shifters are not distributed into all MSA heads and layers but are expanded based on prompt length ($l$). Thus, each element of the tuple, e.g., $a^c_K$ is a vector with $l-1$ elements. Note that the first length $(l=1)$ is not scaled or shifted, but left original $G(x)$. Our method utilizes small prompt length as in DualPrompt \citep{DualP_wang2022dualprompt}, i.e., $l=5$ to maximize the model throughput. Based on that setting, the total scalers and shifters for each class are $4 \times 4 = 16$ parameters. To avoid uncontrollable values of $(a^c_K, b^c_K, a^c_V, b^c_V)$ we deploy upper and lower bounds both for $K$ and $V$ scalers and shifters formulated by the expression below: 
\begin{equation}\label{eq:ss_bound}
    (1-\epsilon_a) \leq a^c \leq (1+\epsilon_a) \text{, and } -\epsilon_b \leq b^c \leq \epsilon_b
\end{equation}
where $\epsilon_a$ and  $\epsilon_b$ are small positive real numbers. The boundaries are deployed to avoid too much perturbation of $G_K(x)$ and $G_V(x)$ that leads to degraded performance. $K^c$ is a $\mathbb{R}^D$ learnable vector that represent class $c$, where $D$ is the embedding size. By representing class $c$, $K^c$ plays as a reference to select the appropriate scalers and shifters during the inference time based on its similarity to input $x$. 

\subsection{Prompt Generation}\label{sec:prompt_gen}
We propose a novel prompt generation mechanism i.e., by single lightweight generators and scalers-and-shifters. The mechanism is unique to existing prompt-generation methods \citep{CODA_smith2023coda,CONVP_roy2024convolutional,EVO_kurniawan2024evolving}.  Given prompt generator $(G_K, G_V)$ and scaler-and-shifters $(a^c_K, b^c_K, a^c_V, b^c_V)$ where each element as $l$-elements vector, the prompt is generated by executing equation \ref{eq:prompt_gen}.
\vspace{-5pt}
\begin{equation}\label{eq:prompt_gen}
    \small
    \begin{split}
        P_K = [P_{K,1};P_{K,2};...;P_{K,l}] \text{, } P_V = [P_{V,1};P_{V,2};...;P_{V,l}] \\
        P_{K,i} = a^c_{K,i-1}G_K(x)+b^c_{K,i-1}, \\
        P_{V,i} = a^c_{V,i-1}G_V(x)+b^c_{V,i-1}, i \in [2..l] 
    \end{split}
\end{equation}
where ";" denotes the concatenation operation. We add similarity factor $s$ to $G(x)$, i.e., $G(x) = s.G(x)$ where $s$ is the cosine similarity between input and closest key, i.e., $s = cosine(x,K^c)$. The addition of the similarity factor acts as a controller if the input chooses the wrong key. Following the previous study\citep{DualP_wang2022dualprompt,CONVP_roy2024convolutional}, the matched input $x$ can be pacthed embedding, class token, or class feature that we set as default in this study. 

\subsection{Current-to-Previous Classes Orthogonality}\label{sec:ortho}
We apply current-to-previous classes orthogonality to increase the discriminative capability of our method. Let the method currently at task $t$ that learns class $j$, $j+1$,$j+2$...$m$. The model must discriminate any currently learned class $c \in [j,j+1,..m]$ to any previously learned class $c' \in [1,2,..j-1]$. Due to rehearsal-free constraints, our model can not replay any samples from class $c'$ with the samples of $c$. Dealing with the constraint, we utilize $K^c$ and $K^{c'}$ that represent $c$ and $c'$ respectively, thus the model can always improve its discrimination capability. Following the vector principle, that states 2 orthogonal vectors have a zero dot product, our method optimizes $K^c$ so that $K^c.K^{c'} = 0$. 

\subsection{Preserving Generalization Property}\label{sec:presenve_gen}
We utilize PTM that has generalization capability as the initial weight of our method. Dealing with the sequence of streaming data that may be infinite, the generalization capability must be preserved so that it can be taken advantage of by the upcoming data. Following the previous study\citep{SAFE_zhao2025safe}, we preserve the generalization capability by cross correlating the feature of PTM (only), i.e., $f_{\theta}(x)$ and the feature of PTM and prompt, i.e., $f_{\theta;P}(x)$ into a matrix $M$ by executing equation \ref{eq:gen_preserve}.
\vspace{-5pt}
\begin{equation}\label{eq:gen_preserve}
    \small M = 1/B [f_{\theta}(x)]^T[f_{\theta;P}(x)]
\end{equation}
M is $D \times D$ matrix, while $f_{\theta}(x)$ and $f_{\theta;P}(x)$ are $B \times D$, where $B$ and $D$ is batch and embedding sizes respectively. The total quantity of the diagonal elements of M shows the alignment between the PTM feature ($f_{\theta}(x)$) and the PTM+prompt tuning feature ($f_{\theta;P}(x)$). The higher the quantity, the higher the generalization preserving. 

\subsection{Learning, Losses, and Final Objectives}\label{sec:learning}
In this subsection, we explain the learning objective of our method. For a better understanding, we explain it in a per-component way as follows.

\noindent \textbf{(a) Tuning Generators and scaler-and-shifters:} These 2 components work together to generate the prompt that is prepended into ViT layers and go through until the last module, i.e., MLP classifier. Along with the MLP classifier, these two components are tuned by using cross entropy loss that is derived into two types, i.e., intra-task loss ($\mathcal{L}_{intra}$) and inter-task loss $\mathcal{L}_{inter}$. $\mathcal{L}_{intra}$ improves model discrimination within a task, specifically, currently learned classes, while $\mathcal{L}_{inter}$  improves model discrimination for all learned classes, including current and old classes. The intra and inter-task loss are formulated as in equations \ref{loss_intra} and \ref{loss_inter}.
\vspace{-5pt}
  \begin{equation}\label{loss_intra}
  \small
    \mathcal{L}_{intra} =  -\sum_{c \in C^t} log \frac{exp(g_{\phi}(f_{\theta;P}(x))[c])}{\sum_{c' \in \mathcal{C}^t} exp(g_{\phi}(f_{\theta;P}(x))[c'])}
 \end{equation}
 \vspace{-5pt}
  \begin{equation}\label{loss_inter}
   \small
    \mathcal{L}_{inter} =  -  \sum_{t=1}^{t=T} \sum_{c \in \mathcal{C}^t} log \frac{exp(g_{\phi}(f_{\theta;P}(x))[c])}{ \sum_{t=1}^{t=T}  \sum_{c' \in \mathcal{C}^t} exp(g_{\phi}(f_{\theta;P}(x))[c'])}
 \end{equation}

 \noindent \textbf{(b) Tuning Task-wise learnable key:} $K^c$ has two important roles in our method, i.e., as the class representation to select scalers-and-shifters and improving model discrimination through orthogonality. As the class representation, we tune $K^c$ tuned closer to the input sample of class $c$ with similarity loss $\mathcal{L}_{sim}$ as expressed in the equation \ref{loss_k}.
  \begin{equation}\label{loss_k}
    \small
    \mathcal{L}_{sim} =  - (x.K^{c})/(max(||x||_2.||K^{c}||_2,\epsilon))
 \end{equation}
To improve class discrimination, we tune any $K^c$ from currently learned classes to be orthogonal to any $K^{c'}$ from previously learned classes. let $y = [c_1,c_2,...c_B]$ is the class labels of batch-wise input $x$, $K_{new} = [K^{c_1}, K^{c_2},...K^{c_B}]$ is the class-wise keys matched with $y$, and  $K_{old} = [K^{c'_1}, K^{c'_2},...K^{c'_B}]$, then the model will be improved by orthogonality loss $\mathcal{L}_{ort}$ as formulated in equation \ref{loss_ort}.
\vspace{-5pt}
  \begin{equation}\label{loss_ort}
  \small
    \mathcal{L}_{ort} =  \frac{1}{B} \sum_{i=1}^{B} K^{ci} . K^{c'i} 
 \end{equation}

 \noindent \textbf{(c) Tuning Generalization preserving Matrix:} 
 The diagonal elements of matrix $M$ represent the agreement between PTM only prototype ($f_{\theta}(x)$) and PTM+Prompt prototype ($f_{\theta;P}(x)$), while the non-diagonal represent the diasagreement between them. Thus we tune the generalization matrix $M$ with generalization loss $\mathcal{L}_{gen}$ as formulated in the equation \ref{loss_gen}. The first term of $\mathcal{L}_{gen}$ push the model towards the agreement, while the second term of the $\mathcal{L}_{gen}$ avoids the disagreement of the prototypes.
 \vspace{-5pt}
 \begin{equation}\label{loss_gen}
 \small
    \mathcal{L}_{gen} =  \frac{1}{D} \sum_{i=1}^{D} (1-M_{i,i})^2 +  \frac{1}{D(D-1)} \sum_{i=1}^{D}\sum_{j \neq i} M_{i,j}^2 
 \end{equation}

 \noindent \textbf{(d) Joint Loss and Final Objective:} 
 Finally, accommodating the losses for all components of our proposed method, i.e., single lightweight generators, scalers-and-shifter, class-wise keys, and generalization matrix,  we define a new joint loss function as in equation \ref{loss_total}. To enhance the flexibility of our model we add losses coefficient $\lambda_1,\lambda_2,\lambda_3,\lambda_4,\lambda_5$. 
 \begin{equation}\label{loss_total}
 \small
 \begin{split}
     \mathcal{L}_{total} = & \lambda_1\mathcal{L}_{intra} + \lambda_2\mathcal{L}_{inter} + \lambda_3\mathcal{L}_{sim} + \\
     & \lambda_4\mathcal{L}_{ort} + \lambda_5\mathcal{L}_{gen}
 \end{split}
 \end{equation}

\noindent \textbf{(e) Hard-Soft Update (HSU) Strategy:} Dealing with streaming data that the learning cannot be repeated into several epochs, we devise a learning strategy for updating the model. We utilize Hard-Soft Update (HSU) where the model adaptively switches from-and-into hard update, i.e., update with constant and high learning rate, and soft update, i.e., update with decayed learning rate. The hard update is started when the model see new class labels and switch into soft update if the cross entropy loss, i.e., $\mathcal{L}_{ce} = \lambda_1\mathcal{L}_{intra} + \lambda_2\mathcal{L}_{inter}$ less than a loss threshold $\mathcal{L}_{thres}$. The decay is executed by a cosine annealing scheduler.
\begin{table*}[h!]
\centering
\footnotesize
\setlength{\tabcolsep}{0.61em}
\caption{Summarized performance of consolidated methods in 4 benchmark dataset, i.e., CIFAR100, Imagnet-R, Imagnet-A, and CUB200. The official code of ConvPrompt is not runnable on Imagenet-A due to its descriptor unavailability to determined new generators for continual tasks. PET and Rhsl denotes parameter efficient tuning and rehearsal respectively.} 
\vspace{-3pt}
\begin{tabular}{llccccccccccccc}
\hline
&  &  & \multicolumn{3}{c}{\textbf{CIFAR100}} & \multicolumn{3}{c}{\textbf{ImageNet-R}} & \multicolumn{3}{c}{\textbf{ImageNet-A}} & \multicolumn{3}{c}{\textbf{CUB200}} \\ \cline{4-15}
\multirow{-2}{*}{\textbf{Method}} & \multirow{-2}{*}{\textbf{PET Type}} & \multirow{-2}{*}{\textbf{Rhsl}} & \textbf{FAA} & \textbf{CAA} & \textbf{FM} & \textbf{FAA} & \textbf{CAA} & \textbf{FM} & \textbf{FAA} & \textbf{CAA} & \textbf{FM} & \textbf{FAA} & \textbf{CAA} & \textbf{FM} \\ \hline
RanPAC-J \citep{RANPAC_mcdonnell2023ranpac} & Rand. matrix & yes & 79.94 & 85.91 & 6.43 & 52.09 & 56.21 & 6.63 & 45.03 & 53.30 & 6.97 & 85.20 & 86.31 & 2.63 \\
RanDumb-J \citep{RANDUMB_prabhu2025randumb} & RBF kernel & yes & 80.39 & 86.15 & 6.25 & 52.63 & 56.13 & 6.04 & 44.07 & 52.30 & 4.98 & 85.82 & 86.84 & 2.52 \\
RanPAC-R \citep{RANPAC_mcdonnell2023ranpac} & Rand. matrix & yes & 59.66 & 69.35 & 34.30 & 28.16 & 35.37 & 39.81 & 28.54 & 35.16 & 32.05 & 75.63 & 79.37 & 14.65 \\
RanDumb-R \citep{RANDUMB_prabhu2025randumb} & RBF kernel & yes & 50.93 & 62.14 & 47.35 & 21.91 & 29.90 & 48.81 & 24.86 & 30.82 & 34.79 & 73.81 & 78.70 & 17.45 \\
RanPAC \citep{RANPAC_mcdonnell2023ranpac} & Rand. matrix & no & 9.52 & 27.92 & 94.52 & 7.24 & 18.81 & 67.27 & 3.77 & 14.86 & 57.76 & 9.52 & 27.06 & 88.98 \\
RanDumb \citep{RANDUMB_prabhu2025randumb} & RBF kernel & no & 9.56 & 27.94 & 94.51 & 7.32 & 18.80 & 67.67 & 3.69 & 14.56 & 55.15 & 9.53 & 27.21 & 89.69 \\
L2P \citep{L2P_wang2022learning} & Prompt (Latent) & no & 80.99 & 86.66 & 6.71 & 52.08 & 55.90 & 5.67 & 14.64 & 21.88 & 5.35 & 61.98 & 71.46 & 9.81 \\
DualPrompt \citep{DualP_wang2022dualprompt} & Prompt (Latent) & no & 82.53 & 88.06 & 4.53 & 61.17 & 62.73 & 3.75 & 20.05 & 21.57 & 1.91 & 58.06 & 49.95 & 6.23 \\
ConvPrompt \citep{CONVP_roy2024convolutional} & Prompt Gen. & no & 84.79 & 88.19 & 2.38 & 70.92 & 74.66 & 3.39 & - & - & - & 70.12 & 77.48 & 7.21 \\
MOS \citep{MOS_sun2024mos} & Adapters & no & 83.99 & 89.53 & 11.34 & 52.13 & 67.97 & 35.72 & 44.79 & 55.14 & 15.50 & 61.74 & 77.03 & 25.25 \\
PROL (Ours) & Prompt. Gen+SS & no & 86.32 & 91.35 & 6.34 & 73.50 & 78.13 & 4.82 & 47.72 & 58.96 & 3.29 & 72.51 & 79.89 & 8.92 \\ \hline
\end{tabular}

\label{tab:4dataset_main_table}
\end{table*}

\begin{figure*}[h!]
\setlength{\abovecaptionskip}{-7pt plus 0pt minus 0pt}
\setlength{\belowcaptionskip}{-10pt plus 0pt minus 0pt}
\begin{center}
\includegraphics[width=0.9\textwidth]{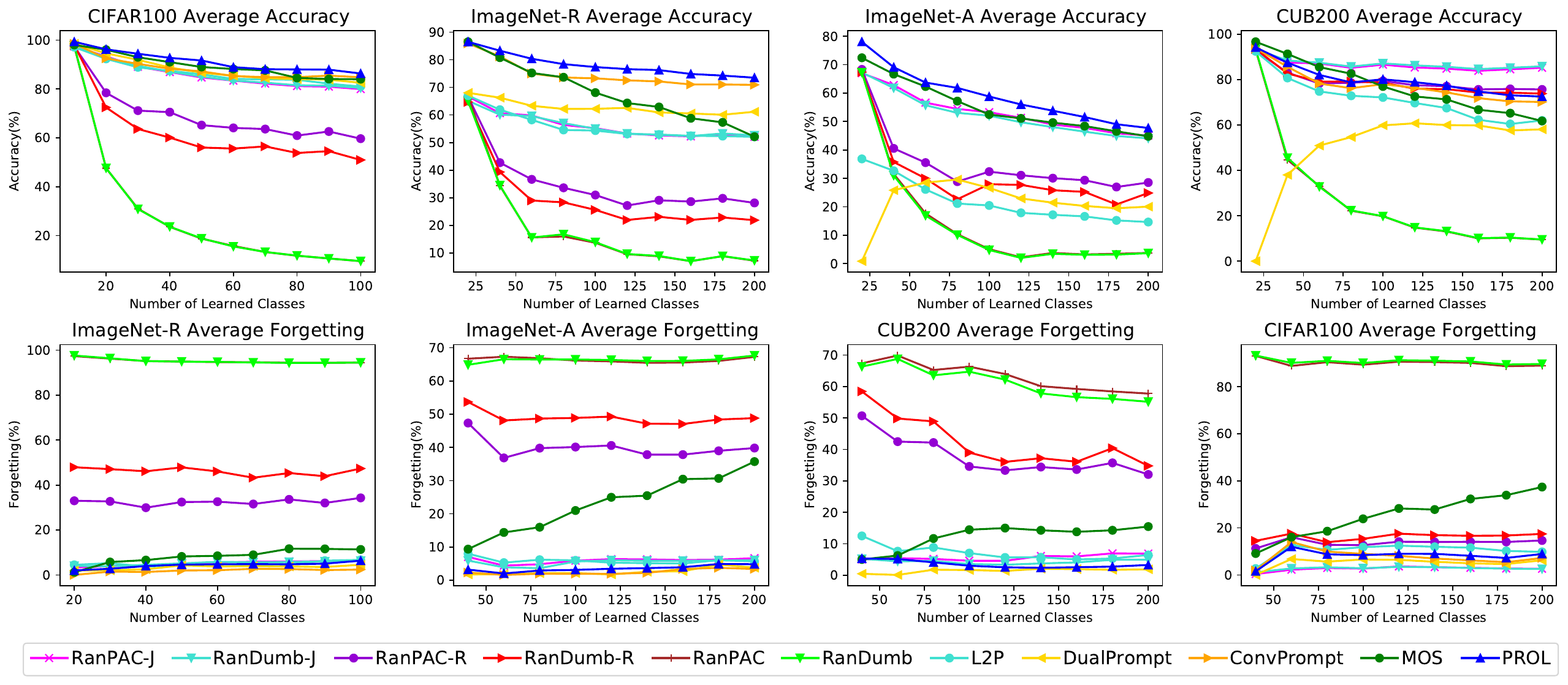}
\end{center}
\caption{Historical performance plot of the consolidated methods in  CIFAR100, Imagnet-R, Imagnet-A, and CUB200 dataset} 
\label{fig:4dataset_plot}
\end{figure*}

\begin{table}[]
\centering
\scriptsize
\caption{Comparison of proposed method to non-PTM method on CIFAR100 dataset, the result is copied from ERSM\citep{ESRM} }
\vspace{-3pt}
\begin{tabular}{lccccc}
\hline
&  & \multicolumn{3}{c}{\textbf{Backbone}} &  \\ \cline{3-5}
\multirow{-2}{*}{\textbf{Method}} & \multirow{-2}{*}{\textbf{Mem}} & \textbf{Type} & \textbf{Size} & \textbf{PTM} & \multirow{-2}{*}{\textbf{\begin{tabular}[c]{@{}c@{}}FAA on\\ CIFAR100\end{tabular}}} \\ \hline
ER \citep{ER_riemer2018learning} & 5000 & ResNet-18 & 11.2M & no & 38.70 \\
DER++ \citep{DER_buzzega2020dark} & 5000 & ResNet-18 & 11.2M & no & 37.62 \\
ERACE \cite{ERASE} & 5000 & ResNet-18 & 11.2M & no & 39.82 \\
OCM \citep{OCM_guo2022online} & 5000 & ResNet-18 & 11.2M & no & 42.01 \\
GSA \citep{GSA_guo2023dealing} & 5000 & ResNet-18 & 11.2M & no & 42.27 \\
OnPro \citep{ONPRO_wei2023online} & 5000 & ResNet-18 & 11.2M & no & 41.47 \\
ERSM \citep{ESRM} & 5000 & ResNet-18 & 11.2M & no & 47.72 \\
PROL (Ours) & - & ViT-Tiny & 5.6M & yes & 64.59 \\ \hline
\end{tabular}
\label{tab:non-ptm}
\vspace{-10pt}
\end{table}

\section{Result and Analysis}
\label{sec:result}
\subsection{Expriment Setting}
 We evaluate our method in 4 of the most popular datasets in CL i.e. CIFAR100\cite{CIFAR100_hendrycks2021many}, ImageNet-R \citep{IMR_krizhevsky2009learning}, ImageNet-A \citep{IMAGENETA_hendrycks2021nae}, and CUB\citep{CUB_wah2011caltech}. CIFAR100 contains 100 classes images, while the other has 200 classes. We uniformly split the dataset into 10 tasks.  We compare our proposed method with the current SOTAs of PTM-based OCL, i.e., RanDumb \citep{RANDUMB_prabhu2025randumb}, RanPAC \citep{RANPAC_mcdonnell2023ranpac}, and MOS \citep{MOS_sun2024mos}, the extension of EASE \citep{EASE_zhou2024expandable}. We adapt the official code of RanDumb and RanPAC from \citep{RANDUMB_prabhu2025randumb} into our rehearsal free and disjoint setting. We also adapt the rehearsal version (denoted with -R) with 5 samples per class as memory, and the joint version (denoted with -J) for both of them.  We adapted MOS from its official code into rehearsal free setting. We add L2P \citep{L2P_wang2022learning}, DualPrompt \citep{DualP_wang2022dualprompt}, and ConvPrompt \citep{CONVP_roy2024convolutional} as representation of pool-based, task-specific, and growing components prompting, respectively. To ensure fairness, all the consolidated methods are run with the same dataset split, backbone, i.e., ViT-B/16 pre-trained on Imagenet-21K, and a 10 chunk size. We also compare our method to non-PTM methods \citep{ER_riemer2018learning,DER_buzzega2020dark,ERASE,OCM_guo2022online,GSA_guo2023dealing,ONPRO_wei2023online,ESRM} as reported in \citep{ESRM}. The learning rates for all methods are set by grid search from the range of [0.001,0.005,0.01,0.05,0.1]. All the methods are run under the same environment, i.e., a single NVIDIA A100 GPU with 40 GB RAM, with 3 different random seeds. We utilize Adam optimizer for PROL, while the rest follow their official implementation. We evaluate final average accuracy (FAA), cumulative average accuracy (CAA), and final forgetting measure (FM), see supplementary. 

\subsection{Result and Analysis}
\vspace{-7pt}
\noindent\textbf{a) Overall Performance:} Table \ref{tab:4dataset_main_table} shows the summarized performance of the consolidated method in 4 benchmark datasets. In term of accuracy, our method outperforms the existing SOTAs with a significant gap, i.e., 2-76\% FAA and 2-64\% CAA, respectively, except in the CUB dataset. Surprisingly, our method outperforms the joint version of RanDumb and RunPAC with a 2-20\% gap both for FAA and CAA, even though those methods join all the samples' features together. Our method outperforms their replay versions (RanPAC-R and RanDumb-R) with a high margin, i.e., 20-35\% FAA and CAA margins. The rehearsal free of those methods perform extremely poor indicated by the very low FAA and CAA. Our method outperforms the prompt-based methods with a significant margin, i.e., 2-20\% margin both for FAA and CAA. ConvPrompt is the closest competitor to our method with 2-3\% difference in all cases. Our method outperforms the adapter method (MOS) with a significant margin, i.e., 2-21\% FAA and 2-10\% CAA, respectively.  In CUB dataset, our method has lower performance than RanPAC-J and RanDumb-J, comparable performance with RanPAC-R and RanDumb-J, and outperforms the other methods with a convincing gap, i.e., 2-63\% FAA and 2-52\% CAA, respectively. In terms of forgetting, our method achieves a fairly low forgetting rate, i.e., slightly higher than Convprompt methods in many cases but significantly lower than RanPAC, RanDumb, and MOS. Our method even achieves lower forgetting in 2 of 4 datasets than RanPAC-J and RanDumb-J.  Table \ref{tab:non-ptm} shows that despite having a smaller backbone size, our method outperform the non-PTM methods with a significant gap, i.e., $\geq16\%$. It shows that PTM and parameter-efficient tuning are preferable to a bigger size backbone and memory.

\noindent\textbf{b) Historical Performance:} Figure \ref{fig:4dataset_plot} plots the detailed performance of the consolidated methods after finishing each task. The figure shows that in CIFAR100, ImageNet-R, and ImageNet-A datasets, our method consistently achieves the highest average accuracy from the first task until the last task. In addition, the figure shows that our method has the smallest slope, indicating that it suffers from the smaller performance degradation in the learning process. In CUB dataset, our method consistently achieves higher performance than all rehearsal-free methods and competes with RanPAC-R and RanDumb-R, even though it has lower performance than RanPAC-J and RanDUmb-J. Note that both the -R and -J methods replay the previous data, and those methods are significantly outperformed by PROL in the other 3 datasets. In terms of the forgetting measure, PROL consistently achieves a fairly low forgetting rate along with the prompt-based methods. The plot shows that our method is consistently in the bottom group while RanPAC, RanDumb, and MOS are in the middle and upper groups.

\begin{figure}[h!]
\centering
\small
\setlength{\abovecaptionskip}{-7pt plus 0pt minus 0pt}
\begin{center}
\includegraphics[width=0.475\textwidth]{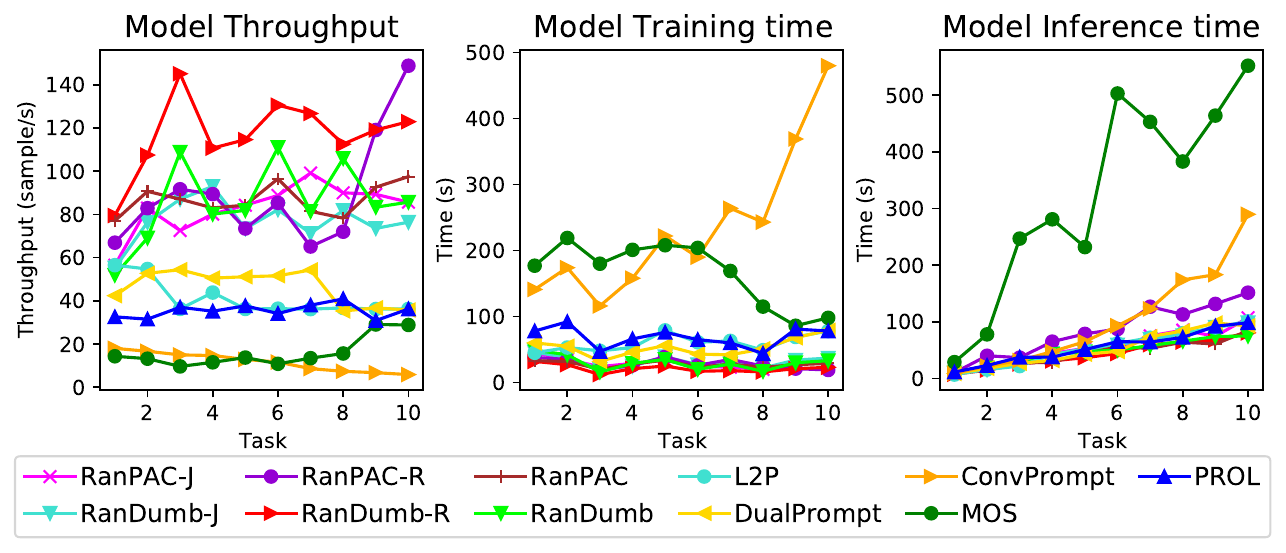}
\end{center}
\caption{Detailed throughput and execution time of consolidated methods in, Imagnet-R dataset.}
\label{fig:throughput}
\end{figure}

\noindent\textbf{c) Parameter, Throughput and Execution time:} Table \ref{tab:throughput} presents the parameter and summarized throughput and execution time while the detailed version is presented by fig.\ref{fig:throughput}. The throughput is computed in the training phase, where the learning process is conducted. The table shows that PROL requires one of the smallest number of trainable parameters, i.e., only slightly bigger than RanPAC and L2P and far smaller than the rest of the methods. RanDumb saves the biggest ($>$ 7M) RBF kernel parameters, while ConvPrompt and MOS require 1.4M and 3.2M parameters, respectively.
The table show that PROL has a moderate throughput, i.e., comparable with L2P and DualPrompt, lower than RanDumb and RanPAC. In line with this, PROL achieves moderate training and inference time. RanDumb and RanPAC have higher throughput and lower execution time since the methods update their feature projection and classifier only once, not iteratively per data stream. In the streaming process, it only extracts the input feature, stores it in memory, and then updates the projection and classifier only once after the current task is finished. The weakness of this mechanism is that it can not handle any time inference where a request for inference comes in the middle of the streaming. 
Figure \ref{fig:throughput} confirms that PROL has moderate throughput, training time, and testing time from the first task until the last task. In terms of throughput, it competes with DualPrompt and L2P and is far better than ConvPrompt and MOS. ConvPrompt and MOS low throughput and high execution time confirm our preliminary analysis on growing prompt components/adapters.

\begin{table}[]
\centering
\scriptsize
\caption{Trainable Parameter (including classifier), summarized throughput and execution time of consolidated methods in, ImageNet-R dataset.}
\vspace{-3pt}
\setlength{\tabcolsep}{0.58em}
\begin{tabular}{lcccc}
\hline
\textbf{Method} & \textbf{\begin{tabular}[c]{@{}c@{}}\#Trainable\\ Parameter\\ In million (M)\end{tabular}} & \textbf{\begin{tabular}[c]{@{}c@{}}Average\\  Throughput\\ (sample/s)\end{tabular}} & \textbf{\begin{tabular}[c]{@{}c@{}}Average \\ Training\\ Time (s)\end{tabular}} & \textbf{\begin{tabular}[c]{@{}c@{}}Average \\ Inference\\ Time (s)\end{tabular}} \\ \hline
RanPAC-J \citep{RANPAC_mcdonnell2023ranpac} & 0.154 & 82.92 & 29.60 & 55.24 \\
RanDumb-J \citep{RANDUMB_prabhu2025randumb} & 7.834 & 77.03 & 32.00 & 53.79 \\
RanPAC-R \citep{RANPAC_mcdonnell2023ranpac} & 0.154 & 89.46 & 28.30 & 84.20 \\
RanDumb-R \citep{RANDUMB_prabhu2025randumb} & 7.834 & 116.89 & 21.20 & 43.36 \\ \hline
RanPAC \citep{RANPAC_mcdonnell2023ranpac} & 0.154 & 86.83 & 27.70 & 45.21 \\ 
RanDumb \citep{RANDUMB_prabhu2025randumb} & 7.834 & 85.90 & 29.90 & 46.81 \\
L2P \citep{L2P_wang2022learning} & 0.200 & 40.94 & 59.80 & 54.50 \\
DualPrompt \citep{DualP_wang2022dualprompt} & 0.701 & 46.50 & 53.20 & 52.60 \\
ConvPrompt \citep{CONVP_roy2024convolutional} & 1.434 & 11.77 & 235.70 & 103.92 \\
MOS \citep{MOS_sun2024mos} & 3.197 & 16.06 & 165.70 & 322.20 \\
PROL (Ours) & 0.213 & 35.42 & 68.70 & 55.39 \\ \hline
\end{tabular}
\label{tab:throughput}
\vspace{-7pt}
\end{table}

\noindent\textbf{d) Robustness Analysis:} Table \ref{tab:robustness} shows the robustness of our method in comparison to the closest competitor, i.e., ConvPrompt, on different prompt lengths and the number of ViT layers pre-pended. The table shows that PROL successfully maintains its performance in different prompt lengths and numbers of layers, even in the smaller prompt length or fewer layers. In addition, it maintains the performance gap to the ConvPrompt as the closest competitor with the convincing margin, i.e. $\geq 2\%$. The table shows that PROL achieves a higher CAA in a higher number of layers. The table also shows that PROL has a better performance stability than ConvPrompt with 2\% performance swing.

\noindent\textbf{e) Ablation Study:} We conducted an ablation study to evaluate the contribution of each component of our method as presented in table \ref{tab:ablation}. The table shows that fine-tuning (FT) alone can not handle OCL problem since the performance is too low. However, combining both inter-task loss $\mathcal{L}_{inter}$ and intra-task loss $\mathcal{L}_{intra}$ has 1-2\% better accuracy than utilizing only one. The generator (G) plays the most important role as it tremendously boosting the accuracy and reduce the forgetting i.e. up to 57\%. The learnable scaler-and-shifetrs (SS) and class-wise keys (K) and input-to-key similarity loss $\mathcal{L}_{sim}$ improve the performance with 1\% margin. At this point, our method is comparable to ConvPrompt as the closest competitor and even achieves higher CAA. Adding orthogonality loss $\mathcal{L}_{ort}$ on K improves both FAA and FAA even though with a small margin. Applying the hard-soft-update strategy improves the model accuracy with 1.5\% FAA and 1.2\% CAA improvement, respectively. However, it increases the forgetting along with the accuracy. Finally, preserving the PTM generalization property by matrix M and generalization loss $\mathcal{L}_{gen}$ improves the accuracy with 1\% margin, and reduces a slight magnitude of forgetting.   

\begin{table}[]
\centering
\scriptsize
\setlength{\tabcolsep}{0.27em}
\caption{Robustness analysis of the proposed method compared to the competitor methods in ImageNet-R dataset.}
\vspace{-3pt}
\begin{tabular}{l|cc|cc|cc|cc|cc}
\hline
\multicolumn{11}{c}{\textbf{Robustness  on Different  Prompt Length}} \\ \hline
 & \multicolumn{2}{c}{\textbf{3}} & \multicolumn{2}{c}{\textbf{5}} & \multicolumn{2}{c}{\textbf{7}} & \multicolumn{2}{c}{\textbf{10}} & \multicolumn{2}{c}{\textbf{15}} \\ \cline{2-11}
\multirow{-2}{*}{\textbf{Method}} & \textbf{FAA} & \textbf{CAA} & \textbf{FAA} & \textbf{CAA} & \textbf{FAA} & \textbf{CAA} & \textbf{FAA} & \textbf{CAA} & \textbf{FAA} & \textbf{CAA} \\ \hline
ConvPrompt & 69.88 & 72.89 & 70.92 & 74.66 & 70.68 & 74.11 & 70.02 & 74.18 & 69.45 & 72.24 \\
PROL & 73.42 & 77.71 & 73.50 & 78.13 & 73.43 & 77.92 & 73.45 & 78.03 & 73.30 & 77.72 \\ \hline
\multicolumn{11}{c}{\textbf{Robustness  on Different  Number of ViT Layers}} \\ \hline
 & \multicolumn{2}{c}{\textbf{3}} & \multicolumn{2}{c}{\textbf{5}} & \multicolumn{2}{c}{\textbf{7}} & \multicolumn{2}{c}{\textbf{9}} & \multicolumn{2}{c}{-} \\ \cline{2-11}
\multirow{-2}{*}{\textbf{Method}} & \textbf{FAA} & \textbf{CAA} & \textbf{FAA} & \textbf{CAA} & \textbf{FAA} & \textbf{CAA} & \textbf{FAA} & \textbf{CAA} & - & - \\ \hline
ConvPrompt & 71.43 & 75.95 & 70.92 & 74.66 & 69.38 & 74.21 & 72.12 & 77.47 & - & - \\
PROL & 73.72 & 77.80 & 73.50 & 78.13 & 73.55 & 78.14 & 73.75 & 78.69 & - & - \\ \hline
\end{tabular}

\label{tab:robustness}
\end{table}

\begin{table}[]
\centering
\scriptsize
\setlength{\tabcolsep}{0.13em}
\caption{Ablation study of our method in ImageNet-R dataset.}
\vspace{-3pt}
\begin{tabular}{llccc}
\hline
\textbf{Component} & \textbf{Loss} & \textbf{FAA} & \textbf{CAA} & \textbf{FM} \\ \hline
FT & $\mathcal{L}_{inter}$ & 11.98 & 26.25 & 75.72 \\
FT & $\mathcal{L}_{inter}$+$\mathcal{L}_{intra}$ & 12.68 & 28.13 & 81.15 \\
FT+G & $\mathcal{L}_{inter}$+$\mathcal{L}_{intra}$ & 69.77 & 74.43 & 3.25 \\
FT+G+SS+K & $\mathcal{L}_{inter}$+$\mathcal{L}_{intra}$+$\mathcal{L}_{sim}$ & 70.82 & 75.83 & 3.96 \\
FT+G+SS+K & $\mathcal{L}_{inter}$+$\mathcal{L}_{intra}$+$\mathcal{L}_{sim}$+$\mathcal{L}_{ort}$ & 71.13 & 76.09 & 4.31 \\
FT+G+SS+K+HSU & $\mathcal{L}_{inter}$+$\mathcal{L}_{intra}$+$\mathcal{L}_{sim}$+$\mathcal{L}_{ort}$ & 72.53 & 77.24 & 4.95 \\
FT+G+SS+K+HSU+M & $\mathcal{L}_{inter}$+$\mathcal{L}_{intra}$+$\mathcal{L}_{sim}$+$\mathcal{L}_{ort}$+$\mathcal{L}_{gen}$ & 73.50 & 78.13 & 4.82 \\ \hline
\end{tabular}
\label{tab:ablation}
\vspace{-7pt}
\end{table}
\section{Concluding Remark}
\vspace{-3pt}
We highlight the rehearsal-free aspect of continual learning in streaming data and propose a novel rehearsal-free method that incorporates four key concepts (1) Single lightweight prompt generator for all instances in all classes (2) learnable scalers-and-shifters parameters associated with learnable class-wise keys (3) cross-correlation matrix for preserving PTM generalization (4) hard-soft update for adaptive learning rate assignment. Our experiment and analysis show the superiority of our method to existing SOTAs in CIFAR100, ImageNet-R, ImageNet-A, and CUB dataset with significant margins, i.e., 2-76\% and 2-64\% margin for FAA and CAA, respectively. Our method significantly outperforms the joint version of existing SOTAs in 3 datasets while running in a rehearsal-free manner. Our historical analysis shows the consistency of our method in maintaining stability-plasticity in every task. Our method achieves moderate training time, and throughput. Our method achieves significantly higher throughput and lower training and testing time than the growing prompt generator and adapters methods. Our robustness analysis shows that our method consistently achieves the highest performance in various prompt lengths and numbers of ViT layers.    

\newpage
\balance
\section*{Acknowledgement}
M. Anwar Ma’sum acknowledges the support of Tokopedia-UI Centre of Excellence for GPU access to run the experiments.

{
    \small
    \bibliographystyle{ieeenat_fullname}
    \bibliography{main}
}

\appendix
\def\theequation{A\arabic{equation}}
\def\thetable{A\arabic{table}}
\def\thefigure{A\arabic{figure}}

\section{PROL Algorithm}
This section presents the PROL training and inference algorithm that is presented in algorithm 1 and 2, respectively.

\section{PROL Complexity Analysis}
This section presents the complexity analysis of our proposed method. Following algorithm 1, the complexity of PROL can be formulated as:

\begin{equation} \label{}
    O(PROL) = \sum_{t=1}^T \sum_{i=1}^{S^t} O(per-stream)
\end{equation}

where $S^t$ is the number of streams in task $t$. Given a stream $s$ on task $t$, PROL executes several operations as presented in lines (5) to (22). All the operations are O(1). Thus PROL complexity can be derived into

\begin{equation} \label{}
    O(PROL) = \sum_{t=1}^T \sum_{i=1}^{S^t} O(|s|)
\end{equation}

Where $|s|$ represents the stream's size. 
Since $\sum_{i=1}^{S^t}|s| = |\mathcal{T}^t|$ then the PROL complexity becomes

\begin{equation} \label{}
    O(PROL) = \sum_{t=1}^T O(|\mathcal{T}|)
\end{equation}

Since $\sum_{t=1}^{T}|\mathcal{T}| = |\mathcal{T}| = N$ then the PROL 
complexity summarized as

\begin{equation} \label{}
    O(PROL) = O(N)
\end{equation}

\begin{algorithm*}[h!]
\caption{PROL Training}\label{alg:leapgen}
\begin{algorithmic}[1]
\State \textbf{Input:} A sequence of tasks $\mathcal{T}^1$, $\mathcal{T}^2$,...,$\mathcal{T}^T$, a frozen  pre-trained ViT $f_{\theta(.)}$ ,
\State Initiate trainable generator $(G_K,G_V)$ and MLP head $g_{\phi}$
\For{$t=1:T$}
    \State stream data $s=\{(x_i,yi)\}$ from $\mathcal{T}^t$
    \State initiate $(K^c, a^c, b^c)$ for any all $c \in s$ if not exist before 
    \If {$t > 1$}
        \State Freeze $(G_K,G_V)$
    \EndIf
    
    \State Generate prompt $(P_K, P_V)$ following eq. (2)
    \State Compute PTM only and PTMP+prompt prototypes: $f_{\theta}(x)$, and  $f_{\theta;P}(x)$
    \State Compute logits:  $g_{\phi}(f_{\theta;P}(x))$
    \State Compute $\mathcal{L}_{intra}$ following eq. (4)
    \State Compute $\mathcal{L}_{inter}$ following eq. (5)
    \State Compute $\mathcal{L}_{sim}$ following eq. (6) 
    \State Compute $\mathcal{L}_{ort}$ following eq. (7) 
    \State Generate $M$ and Compute $\mathcal{L}_{gen}$ as in eq. (8)
    \State Compute $\mathcal{L}_{total}$ following eq. (9)

    \If {$t = 1$}
        \State Update $(G_K,G_V)$ parameters 
    \EndIf
    
    \State update $(K^c, a^c, b^c)$ and $\phi$
    \State Clamp $(a^c, b^c)$ following eq.(1)
    
\EndFor
\State \textbf{Output:}  Optimum generator $(G_K,G_V)$, parameters $(K^c, a^c, b^c)$ and $\phi$ 
\end{algorithmic}
\end{algorithm*}

\begin{algorithm*}
\caption{PROL Inference}\label{alg:leapgen_inf}
\begin{algorithmic}[1]
\State \textbf{Input:} An input $x$, a frozen  pre-trained ViT $f_{\theta(.)}$, optimized generator $(G_K,G_V)$ parameters $(K^c, a^c, b^c)$ and $\phi$

\State Find top-1 $K^c$ where $c \in \mathcal{T}$ 
\State Generate prompt $(P_K, P_V)$ following eq. (2)
\State Compute logits by forwarding the input i.e. $logits = g_{\phi}(f_{\theta; P}(x))$ 
\State Compute Pedicted label $\hat{y} = argmax(logits)$  

\State \textbf{Output:}  Predicted label $\hat{y}$
\end{algorithmic}
\end{algorithm*}

\section{Detailed experimental setting}
 We evaluate our method in 4 of the most popular datasets in CL i.e. CIFAR100\cite{CIFAR100_hendrycks2021many}, ImageNet-R\citep{IMR_krizhevsky2009learning}, ImageNet-A\citep{IMAGENETA_hendrycks2021nae}, and CUB\citep{CUB_wah2011caltech}. CIFAR100 contains 100 classes images, while the other has 200 classes. We uniformly split the dataset into 10 tasks.  We compare our proposed method with the current SOTAs of PTM-based OCL i.e. RanDumb\citep{RANDUMB_prabhu2025randumb}, RanPAC\citep{RANPAC_mcdonnell2023ranpac}, and MOS\citep{MOS_sun2024mos}, the extension of EASE\citep{EASE_zhou2024expandable}. We adapt the official code of RanDumb and RanPAC from \citep{RANDUMB_prabhu2025randumb} into our rehearsal free and disjoint setting. We also adapt the rehearsal version (denoted with -R) with 5 samples per class as memory, and the joint version (denoted with -J) for both of them.  We adapted MOS from its official code into rehearsal free setting. We add L2P \citep{L2P_wang2022learning}, DualPrompt\citep{DualP_wang2022dualprompt}, and ConvPrompt\citep{CONVP_roy2024convolutional} as representation of pool-based, task-specific, and growing components prompting, respectively. To ensure fairness, all the consolidated methods are run with the same dataset split,  backbone i.e. ViT-B/16 pre-trained on Imagenet-21K. We also compare our method to non-PTM methods i.e. ER \citep{ER_riemer2018learning} DER++\citep{DER_buzzega2020dark}, ERASE \citep{ERASE,OCM_guo2022online}, GSA\citep{GSA_guo2023dealing}, ONPRO \citep{ONPRO_wei2023online}, and ERSM\citep{ESRM}. The compared result is taken from \citep{ESRM}. The learning rates for all methods are set by grid search from the range of [0.001,0.005,0.01,0.05,0.1]. All the methods are run under the same environment, i.e., a single NVIDIA A100 GPU with 40 GB RAM by three different random seed numbers. We utilize Adam optimizer for PROL, while the rest follow their official implementation. We evaluate final average accuracy (FAA), cumulative average accuracy (CAA), and final forgetting measure (FM); please see the supplementary document for the details. 

 The $\lambda_1$, $\lambda_3$ and  $\lambda_4$ are set to 1.0, while  $\lambda_2$ is set to 0.001 for CIFAR100, and 0.01 for CUB and 0.03 for ImageNet-R and ImageNet-A. The $\mathcal{L}_{thres}$ is set to 0.3 for CIFAR100 and CUB, and 0.8 for ImageNet-R and ImageNet-A. The cosine annealing scheduler is set to maxT=20 and min-lr=0.005. All the consolidated methods are run under the same machine and computing environment i.e. single NVIDIA A100 GPU with 40 GB memory, python 3.9 and Pytorch 2.2.0.

\noindent\textbf{Performance Metrics:} Adapted from HidePrompt, we measure both accuracy and forgetting of the methods. Suppose that $A_{i,t}$ denotes the accuracy on the $t$-th task after learning the $t$-th task. The average accuracy of all learned task is defined as $AA_t = (1/t) \Sigma_{i=1}^tA_{i,t}$. Suppose that $T$ is the number of all tasks, we measure final average accuracy (FAA), cumulative average accuracy(CAA), and final forgetting measure (FM)
\begin{equation} \label{}
    FAA = AA_T
\end{equation}
\begin{equation} \label{}
    CAA = \frac{1}{T}\Sigma_{t=1}^T AA_t
\end{equation}
\begin{equation} \label{}
    FFM = \frac{1}{T-1}\Sigma_{i=1}^{T-1} max_{t\in {1,...,T-1}} (AA_{i,t}-AA_{i,T})
\end{equation}

\section{Detailed numerical performance}
This section presents detailed numerical results of the consolidated algorithms both accuracy and forgetting, throughput, training time, and inference time as presented in tables \ref{tab:cifar100_acc} to \ref{tab:tstime_imr}
\begin{table*}[h!]
\centering
\footnotesize
\setlength{\tabcolsep}{1.1em}
\caption{Detailed Accuracy performance of consolidated methods in CIFAR100} 
\begin{tabular}{lccccccccccc}
\hline
Method & 1 & 2 & 3 & 4 & 5 & 6 & 7 & 8 & 9 & 10 & AVG \\ \hline
RanPAC-J & 97.40 & 93.32 & 89.10 & 86.91 & 84.70 & 83.40 & 82.18 & 81.17 & 80.94 & 79.94 & 85.91 \\
RanDumb-J & 97.67 & 92.97 & 89.47 & 87.08 & 85.03 & 83.45 & 82.57 & 81.52 & 81.32 & 80.39 & 86.15 \\
RanPAC-R & 97.40 & 78.42 & 71.13 & 70.51 & 65.19 & 64.07 & 63.59 & 60.92 & 62.59 & 59.66 & 69.35 \\
RanDumb-R & 97.90 & 72.42 & 63.62 & 60.08 & 56.03 & 55.61 & 56.48 & 53.80 & 54.55 & 50.93 & 62.14 \\
RanPAC & 97.37 & 47.57 & 31.01 & 23.58 & 18.75 & 15.74 & 13.28 & 11.75 & 10.64 & 9.52 & 27.92 \\
RanDumb & 97.67 & 47.60 & 30.86 & 23.63 & 18.85 & 15.53 & 13.30 & 11.78 & 10.62 & 9.56 & 27.94 \\
L2P & 97.10 & 92.30 & 88.87 & 87.40 & 86.00 & 84.10 & 83.93 & 83.75 & 82.14 & 80.99 & 86.66 \\
DualPrompt & 98.50 & 94.55 & 91.93 & 88.78 & 86.66 & 85.40 & 84.40 & 84.04 & 83.80 & 82.53 & 88.06 \\
ConvPrompt & 98.50 & 92.50 & 90.27 & 88.28 & 87.18 & 85.27 & 84.83 & 84.81 & 85.44 & 84.79 & 88.19 \\
MOS & 98.00 & 96.05 & 92.97 & 90.98 & 88.98 & 88.15 & 87.60 & 84.51 & 84.06 & 83.99 & 89.53 \\
PROL & 99.30 & 96.20 & 94.40 & 92.73 & 91.64 & 88.92 & 88.07 & 88.00 & 87.90 & 86.32 & 91.35 \\ \hline
\end{tabular}

\label{tab:cifar100_acc}
\end{table*}

\begin{table*}[h!]
\centering
\footnotesize
\setlength{\tabcolsep}{1.1em}
\caption{Detailed Forgetting of consolidated methods in CIFAR100} 
\begin{tabular}{lccccccccccc}
\hline
Method & 1 & 2 & 3 & 4 & 5 & 6 & 7 & 8 & 9 & 10 & AVG \\ \hline
RanPAC-J & - & 3.27 & 3.90 & 4.36 & 5.03 & 5.57 & 5.87 & 6.01 & 6.14 & 6.43 & 5.17 \\
RanDumb-J & - & 4.00 & 3.83 & 4.52 & 5.15 & 5.83 & 5.64 & 5.80 & 6.01 & 6.25 & 5.23 \\
RanPAC-R & - & 33.00 & 32.73 & 29.93 & 32.44 & 32.61 & 31.57 & 33.62 & 32.03 & 34.30 & 32.47 \\
RanDumb-R & - & 47.93 & 47.08 & 46.13 & 47.87 & 46.06 & 43.27 & 45.34 & 43.94 & 47.35 & 46.11 \\
RanPAC & - & 97.37 & 96.25 & 95.18 & 94.96 & 94.71 & 94.67 & 94.42 & 94.37 & 94.52 & 95.16 \\
RanDumb & - & 97.63 & 96.42 & 95.13 & 94.94 & 94.81 & 94.54 & 94.37 & 94.38 & 94.51 & 95.19 \\
L2P & - & 4.50 & 5.25 & 3.70 & 4.35 & 4.76 & 4.58 & 4.77 & 6.35 & 6.71 & 5.00 \\
DualPrompt & - & 2.00 & 2.05 & 2.50 & 4.30 & 3.78 & 3.95 & 3.76 & 3.54 & 4.53 & 3.38 \\
ConvPrompt & - & 0.00 & 1.50 & 1.33 & 1.98 & 2.10 & 2.73 & 2.61 & 2.19 & 2.38 & 1.87 \\
MOS & - & 1.20 & 5.75 & 6.60 & 8.23 & 8.50 & 8.95 & 11.66 & 11.60 & 11.34 & 8.20 \\
PROL & - & 2.00 & 2.75 & 3.97 & 4.60 & 4.74 & 4.83 & 4.73 & 5.04 & 6.34 & 4.33 \\ \hline
\end{tabular}
\label{tab:cifar100_for}
\end{table*}

\begin{table*}[h!]
\centering
\footnotesize
\setlength{\tabcolsep}{1.1em}
\caption{Detailed Accuracy of consolidated methods in ImageNet-R} 
\begin{tabular}{lccccccccccc}
\begin{tabular}{lccccccccccc}
\hline
Method & 1 & 2 & 3 & 4 & 5 & 6 & 7 & 8 & 9 & 10 & AVG \\ \hline
RanPAC-J & 66.72 & 60.66 & 59.82 & 56.40 & 55.21 & 53.29 & 52.59 & 52.26 & 53.08 & 52.09 & 56.21 \\
RanDumb-J & 65.12 & 60.11 & 59.61 & 57.06 & 55.02 & 53.27 & 52.79 & 52.37 & 53.31 & 52.63 & 56.13 \\
RanPAC-R & 66.67 & 42.74 & 36.69 & 33.63 & 31.01 & 27.22 & 29.12 & 28.64 & 29.82 & 28.16 & 35.37 \\
RanDumb-R & 64.68 & 39.34 & 29.03 & 28.37 & 25.63 & 21.99 & 23.13 & 22.02 & 22.89 & 21.91 & 29.90 \\
RanPAC & 66.72 & 34.36 & 15.67 & 16.01 & 13.73 & 9.59 & 8.83 & 7.09 & 8.88 & 7.24 & 18.81 \\
RanDumb & 65.12 & 34.53 & 15.62 & 16.80 & 13.95 & 9.68 & 8.96 & 7.07 & 8.91 & 7.32 & 18.80 \\
L2P & 67.01 & 61.92 & 58.23 & 54.58 & 54.35 & 53.13 & 52.93 & 52.44 & 52.35 & 52.08 & 55.90 \\
DualPrompt & 68.02 & 66.26 & 63.33 & 62.17 & 62.23 & 62.57 & 60.92 & 60.54 & 60.08 & 61.17 & 62.73 \\
ConvPrompt & 85.90 & 81.19 & 74.99 & 73.56 & 73.27 & 72.54 & 72.14 & 71.05 & 71.04 & 70.92 & 74.66 \\
MOS & 86.50 & 80.76 & 75.22 & 73.65 & 68.09 & 64.32 & 62.90 & 58.81 & 57.33 & 52.13 & 67.97 \\
PROL & 86.48 & 83.27 & 80.35 & 78.41 & 77.35 & 76.60 & 76.26 & 74.85 & 74.25 & 73.50 & 78.13 \\ \hline
\end{tabular}

\end{tabular}
\label{tab:imr_acc}
\end{table*}

\begin{table*}[h!]
\centering
\footnotesize
\setlength{\tabcolsep}{1.1em}
\caption{Detailed  Forgetting of consolidated methods in ImageNet-R} 
\begin{tabular}{lccccccccccc}
\begin{tabular}{lccccccccccc}
\hline
Method & 1 & 2 & 3 & 4 & 5 & 6 & 7 & 8 & 9 & 10 & AVG \\ \hline
RanPAC-J & - & 7.03 & 4.39 & 4.82 & 5.90 & 6.34 & 6.27 & 6.08 & 6.19 & 6.63 & 5.96 \\
RanDumb-J & - & 5.91 & 3.89 & 3.62 & 5.63 & 6.00 & 5.87 & 5.84 & 5.95 & 6.04 & 5.42 \\
RanPAC-R & - & 47.34 & 36.84 & 39.78 & 40.11 & 40.58 & 37.82 & 37.83 & 38.98 & 39.81 & 39.90 \\
RanDumb-R & - & 53.68 & 48.11 & 48.68 & 48.87 & 49.28 & 47.14 & 47.05 & 48.40 & 48.81 & 48.89 \\
RanPAC & - & 66.72 & 67.31 & 66.86 & 66.16 & 65.90 & 65.47 & 65.55 & 66.05 & 67.27 & 66.36 \\
RanDumb & - & 64.87 & 66.55 & 66.46 & 66.47 & 66.33 & 66.02 & 66.05 & 66.51 & 67.67 & 66.33 \\
L2P & - & 7.99 & 5.30 & 6.18 & 5.99 & 5.26 & 5.14 & 5.04 & 6.04 & 5.67 & 5.85 \\
DualPrompt & - & 1.74 & 1.82 & 2.11 & 2.14 & 1.74 & 2.56 & 2.70 & 4.76 & 3.75 & 2.59 \\
ConvPrompt & - & 2.33 & 1.59 & 1.96 & 1.93 & 1.92 & 2.22 & 3.38 & 3.73 & 3.39 & 2.49 \\
MOS & - & 9.36 & 14.39 & 15.96 & 20.99 & 24.98 & 25.47 & 30.44 & 30.67 & 35.72 & 23.11 \\
PROL & - & 3.20 & 2.03 & 2.93 & 3.11 & 3.45 & 3.65 & 3.90 & 4.89 & 4.82 & 3.55 \\ \hline
\end{tabular}

\end{tabular}
\label{tab:imr_for}
\end{table*}

\begin{table*}[h!]
\centering
\footnotesize
\setlength{\tabcolsep}{1.1em}
\caption{Detailed Accuracy of consolidated methods in ImageNet-A} 
\begin{tabular}{lccccccccccc}
\begin{tabular}{lccccccccccc}
\hline
Method & 1 & 2 & 3 & 4 & 5 & 6 & 7 & 8 & 9 & 10 & AVG \\ \hline
RanPAC-J & 67.00 & 62.80 & 56.63 & 54.42 & 53.27 & 51.29 & 48.93 & 47.74 & 45.86 & 45.03 & 53.30 \\
RanDumb-J & 67.25 & 61.77 & 55.72 & 53.04 & 52.04 & 49.74 & 48.03 & 46.43 & 44.91 & 44.07 & 52.30 \\
RanPAC-R & 68.32 & 40.54 & 35.54 & 28.85 & 32.36 & 31.09 & 30.07 & 29.35 & 26.96 & 28.54 & 35.16 \\
RanDumb-R & 67.37 & 35.76 & 29.98 & 22.64 & 27.95 & 27.71 & 25.83 & 25.24 & 20.81 & 24.86 & 30.82 \\
RanPAC & 67.37 & 31.56 & 17.67 & 10.36 & 5.09 & 2.24 & 3.77 & 3.24 & 3.47 & 3.77 & 14.86 \\
RanDumb & 67.25 & 31.00 & 16.92 & 10.12 & 4.80 & 2.03 & 3.45 & 3.11 & 3.18 & 3.69 & 14.56 \\
L2P & 36.85 & 32.67 & 26.11 & 21.14 & 20.47 & 17.88 & 17.19 & 16.62 & 15.22 & 14.64 & 21.88 \\
DualPrompt & 0.87 & 25.83 & 28.50 & 29.59 & 26.65 & 22.94 & 21.45 & 20.30 & 19.49 & 20.05 & 21.57 \\
ConvPrompt & - & - & - & - & - & - & - & - & - & - & - \\
MOS & 72.41 & 66.67 & 62.30 & 57.17 & 52.45 & 51.09 & 49.60 & 48.39 & 46.53 & 44.79 & 55.14 \\
PROL & 78.04 & 69.09 & 63.70 & 61.82 & 58.80 & 55.99 & 53.80 & 51.57 & 49.03 & 47.72 & 58.96 \\ \hline
\end{tabular}

\end{tabular}
\label{tab:ima_acc}
\end{table*}

\begin{table*}[h!]
\centering
\footnotesize
\setlength{\tabcolsep}{1.1em}
\caption{Detailed  Forgetting of consolidated methods in ImageNet-A} 
\begin{tabular}{lccccccccccc}
\begin{tabular}{lccccccccccc}
\hline
Method & 1 & 2 & 3 & 4 & 5 & 6 & 7 & 8 & 9 & 10 & AVG \\ \hline
RanPAC-J & - & 5.02 & 5.45 & 5.21 & 4.64 & 4.61 & 6.14 & 6.02 & 6.97 & 6.92 & 6.92 \\
RanDumb-J & - & 5.25 & 4.46 & 4.57 & 3.58 & 3.37 & 3.78 & 4.10 & 4.94 & 4.98 & 4.34 \\
RanPAC-R & - & 50.70 & 42.50 & 42.19 & 34.60 & 33.36 & 34.42 & 33.63 & 35.74 & 32.05 & 37.69 \\
RanDumb-R & - & 58.44 & 49.84 & 48.92 & 39.01 & 36.06 & 37.21 & 36.10 & 40.42 & 34.79 & 42.31 \\
RanPAC & - & 67.37 & 69.85 & 65.26 & 66.32 & 63.98 & 60.11 & 59.23 & 58.44 & 57.76 & 63.15 \\
RanDumb & - & 66.34 & 68.77 & 63.56 & 64.72 & 62.20 & 57.82 & 56.62 & 56.07 & 55.15 & 61.25 \\
L2P & - & 12.53 & 7.67 & 8.86 & 7.07 & 5.69 & 5.74 & 5.04 & 5.35 & 6.48 & 62.25 \\
DualPrompt & - & 0.50 & 0.12 & 1.79 & 1.69 & 1.40 & 2.10 & 1.96 & 1.75 & 1.91 & 1.47 \\
ConvPrompt & - & - & - & - & - & - & - & - & - & - & - \\
MOS & - & 5.05 & 6.29 & 11.71 & 14.50 & 15.02 & 14.32 & 13.84 & 14.32 & 15.50 & 12.28 \\
PROL & - & 5.34 & 5.19 & 4.11 & 3.05 & 2.55 & 2.41 & 2.57 & 2.78 & 3.29 & 3.48 \\ \hline
\end{tabular}

\end{tabular}
\label{tab:ima_for}
\end{table*}

\begin{table*}[h!]
\centering
\footnotesize
\setlength{\tabcolsep}{1.1em}
\caption{Detailed Accuracy of consolidated methods in CUB} 
\begin{tabular}{lccccccccccc}
\begin{tabular}{lccccccccccc}
\hline
Method & 1 & 2 & 3 & 4 & 5 & 6 & 7 & 8 & 9 & 10 & AVG \\ \hline
RanPAC-J & 93.11 & 87.48 & 87.10 & 85.17 & 86.54 & 85.32 & 84.81 & 83.84 & 84.51 & 85.20 & 86.31 \\
RanDumb-J & 93.40 & 88.38 & 87.43 & 85.70 & 87.12 & 86.23 & 85.62 & 84.60 & 85.19 & 85.82 & 86.84 \\
RanPAC-R & 92.75 & 83.03 & 78.25 & 78.59 & 79.39 & 77.49 & 77.05 & 75.64 & 75.87 & 75.63 & 79.37 \\
RanDumb-R & 93.46 & 82.66 & 79.08 & 79.17 & 78.58 & 75.99 & 75.74 & 74.27 & 74.20 & 73.81 & 78.70 \\
RanPAC & 93.11 & 44.63 & 33.10 & 22.23 & 19.80 & 14.78 & 13.07 & 10.05 & 10.32 & 9.52 & 27.06 \\
RanDumb & 93.40 & 45.54 & 32.80 & 22.37 & 19.83 & 14.89 & 13.21 & 10.15 & 10.40 & 9.53 & 27.21 \\
L2P & 92.43 & 80.68 & 74.77 & 72.85 & 72.12 & 69.67 & 67.45 & 62.26 & 60.43 & 61.98 & 71.46 \\
DualPrompt & 0.00 & 38.00 & 50.90 & 54.64 & 59.81 & 60.73 & 59.95 & 59.78 & 57.58 & 58.06 & 49.95 \\
ConvPrompt & 93.79 & 85.60 & 78.23 & 76.19 & 77.97 & 76.36 & 74.33 & 71.80 & 70.38 & 70.12 & 77.48 \\
MOS & 96.63 & 91.27 & 85.47 & 82.57 & 76.95 & 72.55 & 71.25 & 66.70 & 65.19 & 61.74 & 77.03 \\
PROL & 94.17 & 87.43 & 82.03 & 78.74 & 80.10 & 78.83 & 77.33 & 74.75 & 73.05 & 72.51 & 79.89 \\ \hline
\end{tabular}

\end{tabular}
\label{tab:cub_acc}
\end{table*}

\begin{table*}[h!]
\centering
\footnotesize
\setlength{\tabcolsep}{1.1em}
\caption{Detailed  Forgetting of consolidated methods in CUB} 
\begin{tabular}{lccccccccccc}
\begin{tabular}{lccccccccccc}
\hline
Method & 1 & 2 & 3 & 4 & 5 & 6 & 7 & 8 & 9 & 10 & AVG \\ \hline
RanPAC-J & - & 0.39 & 2.23 & 2.97 & 2.67 & 3.63 & 3.30 & 3.04 & 2.74 & 2.63 & 2.62 \\
RanDumb-J & - & 1.10 & 2.77 & 3.26 & 2.85 & 3.44 & 3.31 & 2.94 & 2.69 & 2.52 & 2.82 \\
RanPAC-R & - & 11.26 & 16.03 & 12.82 & 12.65 & 14.14 & 14.01 & 14.10 & 14.03 & 14.65 & 13.74 \\
RanDumb-R & - & 14.50 & 17.57 & 13.93 & 15.34 & 17.55 & 16.91 & 16.66 & 16.76 & 17.45 & 16.30 \\
RanPAC & - & 93.11 & 88.86 & 90.45 & 89.38 & 90.66 & 90.48 & 90.13 & 88.75 & 88.98 & 90.09 \\
RanDumb & - & 93.27 & 90.14 & 91.02 & 90.08 & 91.29 & 91.08 & 90.75 & 89.44 & 89.69 & 90.75 \\
L2P & - & 2.72 & 12.68 & 10.66 & 11.81 & 12.45 & 11.97 & 11.60 & 10.22 & 9.81 & 10.44 \\
DualPrompt & - & 0.00 & 6.78 & 5.66 & 6.59 & 6.41 & 5.61 & 4.93 & 4.63 & 6.23 & 5.20 \\
ConvPrompt & - & 1.94 & 13.98 & 9.92 & 8.97 & 8.13 & 6.98 & 6.18 & 5.42 & 7.21 & 7.64 \\
MOS & - & 9.11 & 16.05 & 18.59 & 23.86 & 28.27 & 27.81 & 32.31 & 33.91 & 37.38 & 25.25 \\
PROL & - & 1.55 & 11.98 & 8.77 & 8.43 & 8.98 & 8.97 & 8.13 & 7.22 & 8.92 & 8.11 \\ \hline
\end{tabular}

\end{tabular}
\label{tab:cub_for}
\end{table*}

\begin{table*}[h!]
\centering
\footnotesize
\setlength{\tabcolsep}{1.1em}
\caption{Detailed  Throughput of consolidated methods in ImageNet-R} 
\begin{tabular}{lccccccccccc}
\begin{tabular}{lccccccccccc}
\hline
Method & 1 & 2 & 3 & 4 & 5 & 6 & 7 & 8 & 9 & 10 & AVG \\ \hline
RanPAC-J & 56.51 & 82.89 & 72.54 & 80.14 & 84.26 & 88.76 & 99.13 & 90.00 & 89.29 & 85.67 & 82.92 \\
RanDumb-J & 55.28 & 76.34 & 87.05 & 92.96 & 73.46 & 82.19 & 71.25 & 81.82 & 73.53 & 76.41 & 77.03 \\
RanPAC-R & 66.92 & 82.89 & 91.63 & 89.38 & 73.46 & 85.35 & 65.14 & 72.00 & 119.05 & 148.79 & 89.46 \\
RanDumb-R & 79.47 & 107.44 & 145.08 & 110.67 & 114.60 & 130.53 & 126.67 & 112.50 & 119.05 & 122.91 & 116.89 \\
RanPAC & 77.06 & 90.66 & 87.05 & 83.00 & 84.26 & 96.48 & 81.43 & 78.26 & 92.59 & 97.48 & 86.83 \\
RanDumb & 51.90 & 69.07 & 108.81 & 80.14 & 81.86 & 110.95 & 81.43 & 105.88 & 83.33 & 85.67 & 85.90 \\
L2P & 56.51 & 54.74 & 36.27 & 43.85 & 36.27 & 36.38 & 36.19 & 36.73 & 36.23 & 36.24 & 40.94 \\
DualPrompt & 42.38 & 52.75 & 54.41 & 50.52 & 51.16 & 51.60 & 54.29 & 35.29 & 36.76 & 35.78 & 46.50 \\
ConvPrompt & 18.04 & 16.67 & 15.01 & 14.71 & 12.91 & 11.68 & 8.64 & 7.41 & 6.78 & 5.89 & 11.77 \\
MOS & 14.37 & 13.25 & 9.67 & 11.56 & 13.77 & 10.88 & 13.49 & 15.65 & 29.07 & 28.85 & 16.06 \\
PROL & 32.60 & 31.53 & 37.04 & 35.21 & 37.70 & 34.14 & 38.00 & 40.91 & 30.86 & 36.24 & 35.42 \\ \hline
\end{tabular}

\end{tabular}
\label{tab:tput_imr}
\end{table*}

\begin{table*}[h!]
\centering
\footnotesize
\setlength{\tabcolsep}{1.1em}
\caption{Detailed Training Time of consolidated methods in ImageNet-R} 
\begin{tabular}{lccccccccccc}
\hline
Method & 1 & 2 & 3 & 4 & 5 & 6 & 7 & 8 & 9 & 10 & AVG \\ \hline
RanPAC-J & 45.00 & 35.00 & 24.00 & 29.00 & 34.00 & 25.00 & 23.00 & 20.00 & 28.00 & 33.00 & 29.60 \\
RanDumb-J & 46.00 & 38.00 & 20.00 & 25.00 & 39.00 & 27.00 & 32.00 & 22.00 & 34.00 & 37.00 & 32.00 \\
RanPAC-R & 38.00 & 35.00 & 19.00 & 26.00 & 39.00 & 26.00 & 35.00 & 25.00 & 21.00 & 19.00 & 28.30 \\
RanDumb-R & 32.00 & 27.00 & 12.00 & 21.00 & 25.00 & 17.00 & 18.00 & 16.00 & 21.00 & 23.00 & 21.20 \\
RanPAC & 33.00 & 32.00 & 20.00 & 28.00 & 34.00 & 23.00 & 28.00 & 23.00 & 27.00 & 29.00 & 27.70 \\
RanDumb & 49.00 & 42.00 & 16.00 & 29.00 & 35.00 & 20.00 & 28.00 & 17.00 & 30.00 & 33.00 & 29.90 \\
L2P & 45.00 & 53.00 & 48.00 & 53.00 & 79.00 & 61.00 & 63.00 & 49.00 & 69.00 & 78.00 & 59.80 \\
DualPrompt & 60.00 & 55.00 & 32.00 & 46.00 & 56.00 & 43.00 & 42.00 & 51.00 & 68.00 & 79.00 & 53.20 \\
ConvPrompt & 141.00 & 174.00 & 116.00 & 158.00 & 222.00 & 190.00 & 264.00 & 243.00 & 369.00 & 480.00 & 235.70 \\
MOS & 177.00 & 219.00 & 180.00 & 201.00 & 208.00 & 204.00 & 169.00 & 115.00 & 86.00 & 98.00 & 165.70 \\
PROL & 78.00 & 92.00 & 47.00 & 66.00 & 76.00 & 65.00 & 60.00 & 44.00 & 81.00 & 78.00 & 68.70 \\ \hline

\end{tabular}
\label{tab:trtime_imr}
\end{table*}

\begin{table*}[h!]
\centering
\footnotesize
\setlength{\tabcolsep}{1.1em}
\caption{Detailed Inferfence Time of consolidated methods in ImageNet-R} 
\begin{tabular}{lccccccccccc}
\begin{tabular}{lccccccccccc}
\hline
Method & 1 & 2 & 3 & 4 & 5 & 6 & 7 & 8 & 9 & 10 & AVG \\ \hline
RanPAC-J & 10.29 & 21.00 & 27.37 & 44.49 & 53.27 & 53.38 & 75.35 & 85.34 & 74.61 & 107.30 & 55.24 \\
RanDumb-J & 9.70 & 15.80 & 28.27 & 44.34 & 45.73 & 59.51 & 69.59 & 75.77 & 90.66 & 98.53 & 53.79 \\
RanPAC-R & 10.34 & 40.23 & 36.92 & 65.13 & 78.80 & 87.39 & 126.79 & 113.13 & 131.69 & 151.57 & 84.20 \\
RanDumb-R & 7.74 & 14.80 & 25.75 & 31.27 & 36.84 & 44.17 & 59.77 & 64.45 & 68.29 & 80.57 & 43.36 \\
RanPAC & 10.89 & 22.25 & 27.39 & 28.72 & 44.23 & 55.75 & 53.36 & 64.28 & 60.94 & 84.30 & 45.21 \\
RanDumb & 9.97 & 18.95 & 25.77 & 42.52 & 47.30 & 50.34 & 58.10 & 65.96 & 74.19 & 75.05 & 46.81 \\
L2P & 7.00 & 16.00 & 22.00 & 42.00 & 53.00 & 62.00 & 72.00 & 80.00 & 90.00 & 101.00 & 54.50 \\
DualPrompt & 9.00 & 18.00 & 27.00 & 32.00 & 43.00 & 48.00 & 74.00 & 84.00 & 98.00 & 93.00 & 52.60 \\
ConvPrompt & 9.27 & 22.18 & 31.75 & 47.56 & 65.13 & 93.09 & 122.91 & 174.15 & 183.34 & 289.80 & 103.92 \\
MOS & 29.00 & 78.00 & 247.00 & 281.00 & 232.00 & 503.00 & 453.00 & 383.00 & 464.00 & 552.00 & 322.20 \\
PROL & 11.46 & 23.07 & 37.61 & 37.77 & 51.12 & 65.09 & 64.43 & 72.62 & 92.26 & 98.52 & 55.39
\end{tabular} \\ \hline

\end{tabular}
\label{tab:tstime_imr}
\end{table*}

\end{document}